\documentclass[journal]{IEEEtran}

\usepackage{booktabs} 
\usepackage{url}
\usepackage{multirow}
\usepackage{multicol}
\usepackage{subfigure}
\usepackage{xcolor}
\usepackage{color}
\usepackage{flushend}
\usepackage{array}
\usepackage{todonotes}
\usepackage{cite}
\usepackage{amsbsy}
\usepackage{amsthm}
\usepackage{amsmath}
\usepackage{amsfonts}
\usepackage{graphicx}
\usepackage{enumitem}
\setlist[itemize]{leftmargin=*}

\begin{document}
\author{Yujuan Ding*, Yunshan Ma*, Lizi Liao, Wai Keung Wong,~and Tat-Seng~Chua
\thanks{This research is supported by the National Research Foundation, Singapore under its International Research Centres in Singapore Funding Initiative. Any opinions, findings and conclusions or recommendations expressed in this material are those of the author(s) and do not reflect the views of National Research Foundation, Singapore. We also appreciate the fashion recognition API service provided by Visenze.}
\thanks{Y. Ding is with the College of Computer Science and Software Engineering, Shen zhen University, Shenzhen, 518060, China (e-mail: dingyujuan385@gmail.com)}
\thanks{W. K. Wong is with the Hong Kong Polytechnic University, Hong Kong SAR, China, he is also with the Laboratory for Artificial Intelligence in Design, Hong Kong SAR, China ( calvin.wong@polyu.edu.hk).}
\thanks{Y. Ma, L. Liao, and T. Chua are with National University of Singapore, Singapore (e-mail: yunshan.ma@u.nus.edu, liaolizi.llz@gmail.com, and dcscts@nus.edu.sg)}
\thanks{The corresponding author is W. K. Wong}
\thanks{* The two authors contribute equally.}
}

\title{Leveraging Multiple Relations for Fashion Trend Forecasting Based on Social Media}

\maketitle

\begin{abstract}
Fashion trend forecasting is of great research significance in providing useful suggestions for both fashion companies and fashion lovers. Although various studies have been devoted to tackling this challenging task, they only studied limited fashion elements with highly seasonal or simple patterns, which could hardly reveal the real complex fashion trends. Moreover, the mainstream solutions for this task are still statistical-based and solely focus on time-series data modeling, which limit the forecast accuracy. Towards insightful fashion trend forecasting, previous work~\cite{ma2020knowledge} proposed to analyze more fine-grained fashion elements which can 
informatively reveal fashion trends. Specifically, it focused on detailed fashion element trend forecasting for specific user groups based on social media data. In addition, it proposed a neural network-based method, namely KERN, to address the problem of fashion trend modeling and forecasting. In this work, to extend the previous work~\cite{ma2020knowledge}, we propose an improved model named Relation Enhanced Attention Recurrent (REAR) network. Compared to KERN, the REAR model leverages not only the relations among fashion elements, but also those among user groups, thus capturing more types of correlations among various fashion trends. To further improve the performance of long-range trend forecasting, the REAR method devises a sliding temporal attention mechanism, which is able to capture temporal patterns on future horizons better. 
Extensive experiments and more analysis have been conducted on the FIT~\cite{ma2020knowledge} and GeoStyle~\cite{mall2019geostyle} datasets to evaluate the performance of REAR. Experimental and analytical results demonstrate the effectiveness of the proposed REAR model in fashion trend forecasting, which also show the improvement of REAR compared to the KERN. 
\end{abstract}


\IEEEpeerreviewmaketitle

\section{Introduction}
\IEEEPARstart{F}{ashion} trend forecasting, aiming to master the changeability in fashion, is of great significance and matters as much as it ever did. Although rapid technological change has infiltrated every aspect of modern life, people's wants to convey a sense of self through their appearance have not changed. They need a fashion guidance to develop good taste and catch up with trend~\cite{brannon2005fashion,yang2019interpretable}. Additionally, for the fashion industry, valid forecasting enables fashion companies to establish marketing strategies wisely and continuously anticipate and fulfill their consumers' wants and needs. Traditionally, fashion experts need to travel and conduct surveys to obtain people's real fashion tastes based on local culture and tradition, which usually affects the world's fashion trends~\cite{kim2021fashion}. However, such approaches are inefficient, expensive, and highly dependent on the experts' background, which usually introduces bias to the forecasting results~\cite{dubreuil2020traditional}. Meanwhile, the huge progress in Internet, Big Data, and Artificial Intelligence provides us with an alternative way to tackle this challenging problem: automatic forecasting of fashion trends based on fashion data~\cite{israeli2017predicting}.

\begin{figure}[!tp]
	\centering
	\includegraphics[scale = 0.3]{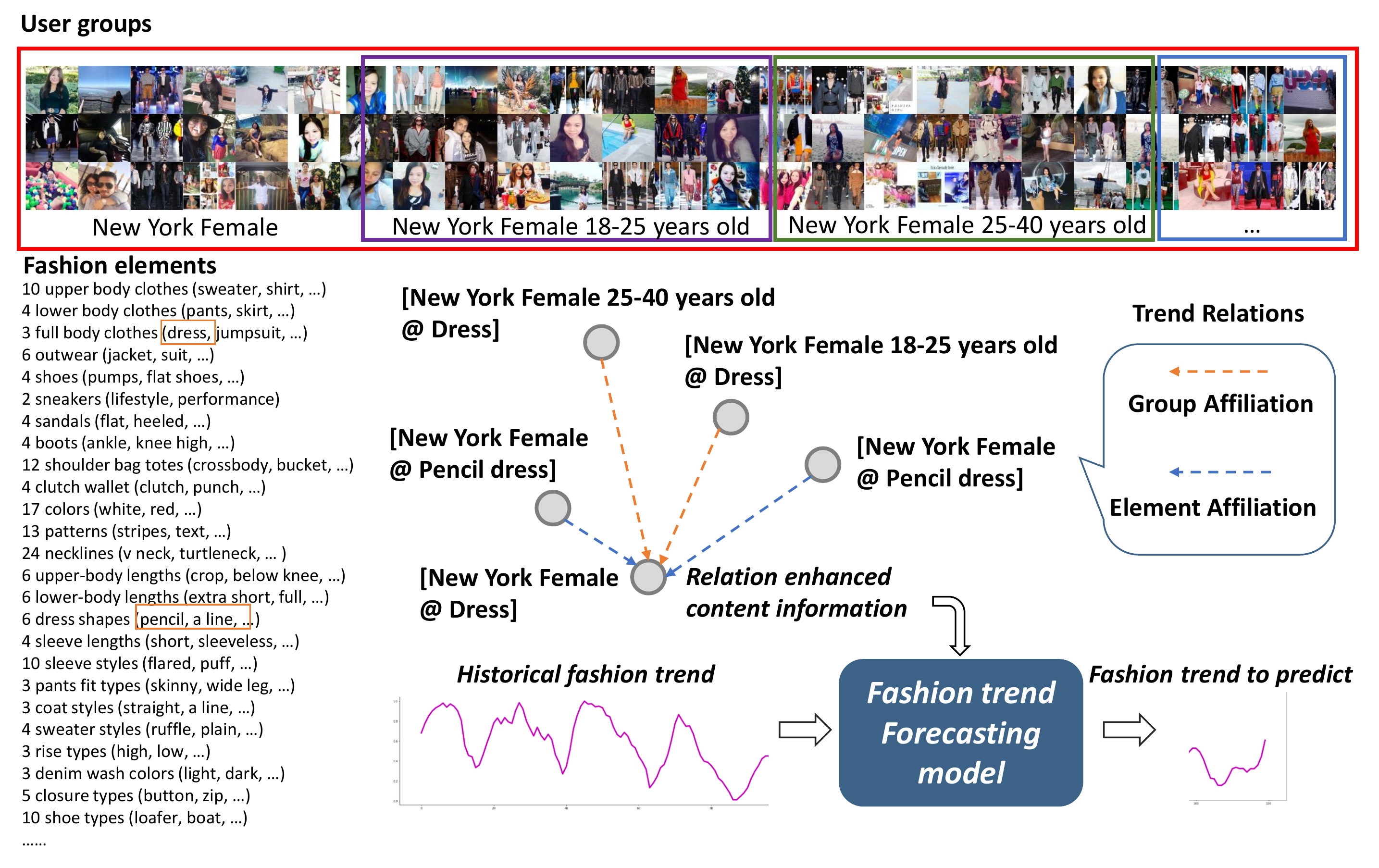}
	\caption{The fashion trend forecasting task aims to forecast the trends of meaningful fashion elements for specific user groups. Multiple relations (group and fashion element affiliations) among the fashion trend time series are leveraged to enhance the forecasting model.
	}
	\label{Fig:task}
\end{figure}

In fact, fashion trend forecasting or analysis has attracted some research interests in the computing field, but is still at its nascent stage. One popular data source for fashion trend analysis is the purchase record on either online or offline retailing platform~\cite{al2017fashion}. However, purchase records reflect people's buying decisions which are influenced by many factors such as retailers' promotions, which hardly reveal real fashion preferences of users, neither the real fashion trend. 
Social media records the daily life of people from all over the world and has turned into a platform for more and more users to show their fashion tastes and opinions. It thus offers a natural platform for research on fashion trend analysis~\cite{ma2019and}. Moreover, data from social media is massive, diverse, highly related to fashion trend and with long time-span, which makes it possible and applicable for insightful large-scale fashion trend analysis. One recent state-of-the-art study of fashion trends based on social media~\cite{mall2019geostyle} extracted fashion attributes from social media images with a CNN model and investigated the trends of specific attributes in certain cities. Compared with other works which target on implicit fashion styles by visual clustering~\cite{matzen2017streetstyle,al2020paris}, the fashion trends demonstrated in Mall's work were more specific. However, the fashion attributes it investigated to reveal specific fashion trends were coarse-grained and of less significance, sometimes showing seasonal patterns only, for example, \textit{wearing jacket}. Moreover, the attributes of users were not explored sufficiently. It involved location information of users while overlooked other basic attributes such as age ad genders. 

Ma \textit{et al.} made an improvement later by investigating fashion trends of specific fashion elements for various user groups. They focused on more fine-grained fashion elements, incorporated more user profile information, and worked on longer fashion trends. To handle the challenging task, a model named KERN that based on an LSTM encoder-decoder framework was proposed. The novel model leveraged the internal and external knowledge to utilize the correlations among specific fashion trend signals to enhance the trend forecasting.

In this paper, we extend the work of \cite{ma2020knowledge}, improve the KERN model, and provide more analysis and insights based on extensive experiments. By analyzing the experimental results reported in~\cite{ma2020knowledge} (specifically in Table 2~\cite{ma2020knowledge}) and our reproduction, we discover that leveraging internal knowledge is not as helpful as external knowledge in KERN. As a result, in this paper, we abandon the internal knowledge and focus on the external knowledge solely. To investigate further on the external knowledge incorporation, we propose to leverage two types of external relations among fashion trends from two perspectives, the fashion element relations, which have been adopted by KERN, and user group relations. Al-Halah \textit{et al.}~\cite{al2020paris} explore fashion influences between different cities, the conclusions of which coincide and support our hypothesis to a certain extent that the fashion trends of different user groups are not independent as well. Specifically in our task, it is natural to consider that the trends of any fashion element for a user group (\textit{e.g.}, \textit{New York Female}) are probably correlated with the corresponding trends of its subset (affiliated) group (\textit{e.g.}, \textit{New York Female 25-40}). As shown in Fig.~\ref{Fig:task}, each fashion trend should be connected to others based on the affiliation relations among the user groups and fashion elements.

Furthermore, as the fashion trend forecasting task aims to predict the trend for a long period, which is more challenging than the one-step-ahead prediction in most time-series forecasting tasks. To effectively capture the temporal patterns on future horizons and enhance the long-period forecasting, we equip our model with a sliding temporal attention module~\cite{fan2019multi}. Specifically, at each time step we make the decoder hidden state to attend to several different periods of the history and generate the attention vector individually. It then combines attention results of all period of the history to generate new representation of the current decoder hidden state. Therefore, the combined features can better describe the current time step as it incorporates both historical information and future contextual information. With above technical improvement, a novel method \textbf{R}elation \textbf{E}nhanced \textbf{A}ttention \textbf{R}ecurrent network (\textbf{REAR}) is proposed in this paper.

This paper is an extension of \cite{ma2020knowledge}, we summarize the contributions of it from two aspects as follows:

\noindent 1) we propose a novel relation-enhanced attention recurrent network (REAR) method for tackling the fashion trend forecasting task. Technically, we improve the KERN~\cite{ma2020knowledge} from two perspectives. First, we leverage more external fashion trend relations. On top of the fashion element relations, the user group relations are also leveraged to bring more association of fashion trend signals. Both relations are incorporated by the message passing modules. Second, to further enhance the performance of long-horizon forecasting, we devise a sliding temporal attention mechanism to effectively capture the temporal patterns on future horizons. 

\noindent 2) Extensive experiments are conducted on the FIT and the GeoStyle datasets~\cite{mall2019geostyle} to evaluate the effectiveness of the proposed REAR method. Specifically, REAR shows superior performance in fashion trend forecasting in terms of overall forecasting accuracy compared to all other competitive baselines, including the KERN. The technical parts designed in REAR are evaluated to be effective via ablation studies. Some technical details that were same as KERN but not investigated in the previous work~\cite{ma2020knowledge} are also discussed in this paper. Furthermore, more insightful analysis and discussion are provided in this paper, including the visualization of the fashion trend forecasting results, the analysis on the influence of multiple relations, and the comparison of specific forecasting results from our REAR model and that from professional fashion trend forecasting agency.

\section{related work}
\subsection{Research targets and datasets for fashion trend forecasting}
Fashion trend forecasting, as an up-stream research task in the field of computational fashion analysis, is attracting increasing research focus nowadays. It is usually studied based on some fundamental related tasks such as fashion recognition, detection, retrieval, and segmentation~\cite{ding2018fashion, liu2016deepfashion, li2017mining, wang2018attentive, ding2019bilinear, gu2018multi, liu2013fashion,ding2019study}. For example, Simo-Serra~\cite{simo2015neuroaesthetics} studied the semantic outfit descriptions based on (non-visual) clothing meta-data, and mainly consists of color and coarse categories. They proposed the Fashion144k dataset for the research based on the fashion website \textit{chictopia} which mainly explored meta-data. The descriptive words in the dataset to determine fashion trends were very sparse and less meaningful to some extent because they were edited casually by users. Al-Halah \textit{et al.}~\cite{al2017fashion} studied fashion trends based on fashion styles. To obtain fashion styles, they represented all fashion images with detected semantic attributes through a pre-trained deep attribute detection model and then applied non-negative matrix factorization on all predicted attributes. They built a dataset based on the e-commence platform Amazon to support their research. The dataset consists of images and text of the purchased items for a fairly long period of time. First, the research target they chose, i.e., the fashion style, is implicit and actually obtained by the clustering of images conditioned with attributes. It thus lacks explanations, which causes the trend analysis to be less effective and convincing. Second, the e-commence data is not suitable for fashion trend analysis because the purchase decision is actually affected by many factors, and fashionability of the item is only one of them. Another very recent research~\cite{al2020paris} also adopted similar strategy. It trained a neural network model to detect the fashion attributes and learned a set of fashion styles based on attributes through Gaussian mixture model. The dataset they adopted is the GeoStyle dataset proposed by Matzen \textit{et al.}~\cite{matzen2017streetstyle, mall2019geostyle}. In this dataset, more specific fashion attributes such as the neckline shape or sleeve length were explored. However, although the attributes they investigated were more detailed, they were of less significance in terms of indicating real fashion and effectively revealing fashion trends. For instance, one fashion element in the dataset was \textit{wearing jacket}, which apparently did not show the evolution of fashion, but the change of seasons. In summary, existing fashion trend forecasting datasets are generally limited and cannot support the study of practical fashion trend and forecast. Towards this research gap, Ma \textit{et al.}~\cite{ma2020knowledge} proposed a large-scale fashion trend dataset based on Instagram (FIT), showing very specific, detailed and realistic fashion trends with respect to a large number of fine-grained fashion elements. It also contained rich user information so that the fashion trends among specific group of people could be further analyzed. The FIT dataset is therefore applied in this work for the further investigation of insightful data-driven fashion trend forecasting. 

\subsection{Time series prediction for fashion trend forecasting}
Fashion trend forecasting aims to predict the future value based on historical time series records. It can be subsumed to the time series prediction problem. Here we review classic and state-of-the-art time series prediction solutions. Autoregrassive (AR)~\cite{walker1931periodicity} is a traditional, simple yet effective statistical model to address the time series prediction problem. Based on that, several advanced models are developed and widely used to solving similar tasks, including moving averages (MA)~\cite{slutzky1937summation}, improved autoregressive integrated moving average (ARIMA)~\cite{box1968some}, and others~\cite{holt2004forecasting,winters1960forecasting}. Although statistical models have achieved great success, they are limited to modeling quite simple or cyclic patterns. In many real life applications, such as the fashion trend forecasting in this paper, the data patterns are notorious for being highly volatile and complex to be captured by statistical models. 

Recently, Neural Networks (NNs) have gradually become powerful techniques for many essential tasks. Specifically, Recurrent Neural Networks (RNNs), especially its variant Long-Short-Term-Memory (LSTM)~\cite{hochreiter1997long}, have achieved state-of-the-art performance in time series prediction~\cite{fan2019multi,liang2018geoman,qin2017dual, zhao2017videowhisper} and been successfully applied in many specific tasks such as stock prediction~\cite{feng2019temporal} and sales forecasting~\cite{bandara2019sales}. Even though fashion trend forecasting is also a type of time series prediction tasks, it is more than that. As a domain-specific task, good solutions should be able to leverage specific and beneficial knowledge in the specific domain. Such idea has also been proposed and implemented in other time series prediction task. In stock prediction, Feng \textit{et al}.~\cite{feng2019temporal} incorporated domain knowledge of stocks and effectively improved the stock price forecasting. However, in fashion trend forecasting, limited attempts have been made. One representative work is from Ma \textit{et al.}~\cite{ma2020knowledge}, which proposed an LSTM-based fashion trend forecasting model KERN. The model is proved effective to make solid multi-horizon trend forecasting. However, it still has technical limitations which has been introduced in the Introduction section. In this paper, a new advanced model is proposed to extend the work of Ma \textit{et al.} and improve the fashion trend forecasting performance based on the KERN model. The new model improves the KERN from three main perspectives. 1) removing the internal knowledge being incorporated which has been shown not to be sufficiently effective through our reproduced experiments; 2) incorporating more external relations; and 3) introducing a sliding attention window mechanism which further improve the fashion trend forecasting performance. 

\section{Problem formulation and dataset}~\label{problem_dataset}
The problem is formulated similar as in \cite{ma2020knowledge}, which aims to make prediction of future popularity with regard to each fashion element for each user group. Given a fashion element $f\in \mathcal{F}$ and a user group $g\in \mathcal{G}$, the temporal popularity of $f$ for $g$ is defined as a time series denoted as $\pmb{y}_g^f=(y_1, \cdots, y_t, \cdots)$, where $\mathcal{F}$ is the set of all fashion elements; and $\mathcal{G}$ is the set of all user groups. The value of the time series at each time step $t$ is defined as $y_t=N_t^{g,f}/N_t^{g}$, where $N_t^{g,f}$ is the number of the fashion elements $f$ observed at time point $t$ for group $g$; $N_t^{g}$ is the number of all fashion items (\textit{e.g, clothing, bags, shoes, and etc.}) observed at time point $t$ for group $g$. Given the historical inputs within the time span of $[1, T]$, our aim is to forecast the future values of time $[T+1, T+T']$, where $T$ is the historical sequence length or time span, and $T'$ is termed as the forecast horizon (the number of steps ahead to forecast, $T'>1$).

\renewcommand\arraystretch{1.3}
\begin{table}
  \caption{Statistical Comparison between FIT and GeoStyle datasets}
  \begin{center}
  \label{tab:statistics}
  \begin{tabular}{p{7.5mm}p{2.5mm}p{6mm}p{5mm}p{10mm}p{8mm}p{5mm}p{7mm}}
    \hline
    Dataset & City &Gender &Age group & Fashion Element & Time span & Data Point /Seq & Sequence\\
    \cline{1-8} 
    \text{GeoStyle} &44 &N/A &N/A &46 &3 years & 144 &2024\\
    \text{FIT} &14 &2 &4 &173 &5 years & 120 &8808\\
    \hline
  \end{tabular}
  \end{center}
\end{table}

\begin{figure}[!tp]
	\centering
	\includegraphics[scale = 0.55]{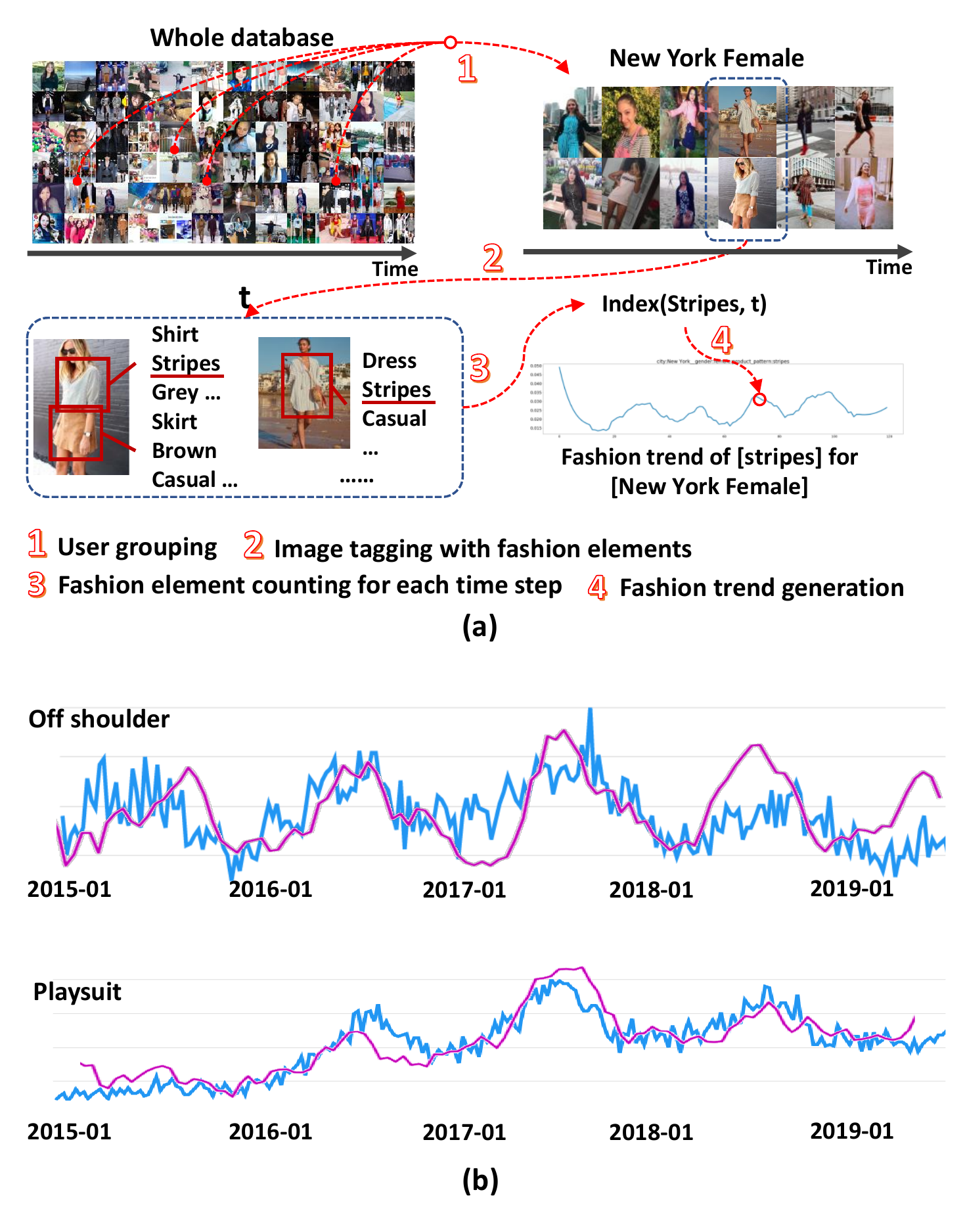}	
	\vspace{-0.05in}
	\caption{(a) Illustration of generating a piece of fashion trend in FIT dataset. (b) Two examples of the FIT dataset, where PURPLE curves are from the FIT dataset and BLUE curves are from Google Trends (both examples belong to the group [New York, Female]).}
	\label{Fig:dataset}
	\vspace{-0.15in}
\end{figure}

Two datasets, FIT~\cite{ma2020knowledge} and GeoStyle~\cite{mall2019geostyle}, are applied in this paper to evaluate different methods for the target problem. Both datasets are based on social media data  and consist of fashion trend signals for long period of time for specific fashion elements and groups of people. Table~\ref{tab:statistics} gives the statistical comparison between FIT and the Geostyle~\cite{mall2019geostyle}. From the comparison, we can see that FIT dataset has more user information, richer fashion elements, and longer time span, thus we take FIT as the main dataset to evaluate most technical parts in our proposed method. More details of the two datasets are introduced as follows.

\textbf{GeoStyle} is adapted from two datasets, StreetStyle~\cite{matzen2017streetstyle} based on Instagram and Flickr 100M dataset~\cite{thomee2016yfcc100m} based on Flickr. It firstly groups all users into 44 major cities, including \textit{Austin}, \textit{Rome}, \textit{Los Angles}, etc. No other user information is provided except for cities (locations). From the perspective of fashion attributes, GeoStyle applies a multi-task learning CNN to train and annotate the images from users. For the annotations, there are 14 binary attributes (\textit{wearing Jacket}, \textit{wearing scarf}, \textit{collar presence}, etc), 13 colors (\textit{black}, \textit{white}, etc), 7 categories (\textit{shirt}, \textit{outwear}, etc), 3 sleeve length (\textit{long}, \textit{short}, \textit{no sleeve}), 3 neckline shape (\textit{round}, \textit{folded}, \textit{v-shape}) and 6 clothing pattern (\textit{solid}, \textit{graphics}, etc), 46 different fashion elements in total. Readers can refer to~\cite{matzen2017streetstyle} for more details about the fashion attributes adopted by GeoStyle. The temporal trend for each attribute in each city for each week is generated by computing the average appearances proportion of an attribute across all photos from that week and city~\cite{mall2019geostyle}. 

\textbf{FIT} is a fashion trend forecasting dataset recently proposed based on Instagram, which originally consists of millions of posts from users. Three types of user information are included in FIT (\textit{i.e.}, age, gender and location). Please refer to \cite{ma2020knowledge} for the details of user information extraction in FIT. Briefly, user information includes 14 main cities (\textit{Paris}, \textit{New York}, \textit{London}, etc), 4 age groups (\textit{18-}, \textit{18-25}, \textit{25-40}, \textit{40+}) and 2 genders (\textit{female} and \textit{male}). Users are separated into different groups based on the user information, resulting in 74 groups of the whole dataset eventually. FIT applies a commercial fashion tagging tool\footnote{visenze.com} to extract three types of fashion elements (category, attribute and style) from the images, resulting in a total of 173 different fashion elements for the whole dataset, much more than GeoStyle.
The popularity of each fashion element for each user group is calculated for every half month as the trend index of the specific period. The post time of FIT dataset ranges from July 2014 to June 2019, spanning five years, which means that each time series has 120 data points. There are 8808 pieces of time series data in FIT in total, excluding sparse series which have less than \textit{50\%} of valid data points. 
The specific process of FIT to generate one piece of fashion trend is shown in Fig.~\ref{Fig:dataset}. We also show the cases of comparison between the fashion trends in FIT and those from \textbf{Google Trends}\footnote{trends.google.com}. They validate the credibility of our FIT dataset (see examples in Fig.~\ref{Fig:dataset}). It should be noted that we use Google Trends to justify the reliability of our dataset; however, Google Trends can be only used for certain popularly searched rough-level fashion elements, it is unable to provide trends of most fine-grained fashion elements and for specific groups of users.

\begin{figure*}[tp]
	\centering
	\includegraphics[scale=0.5]{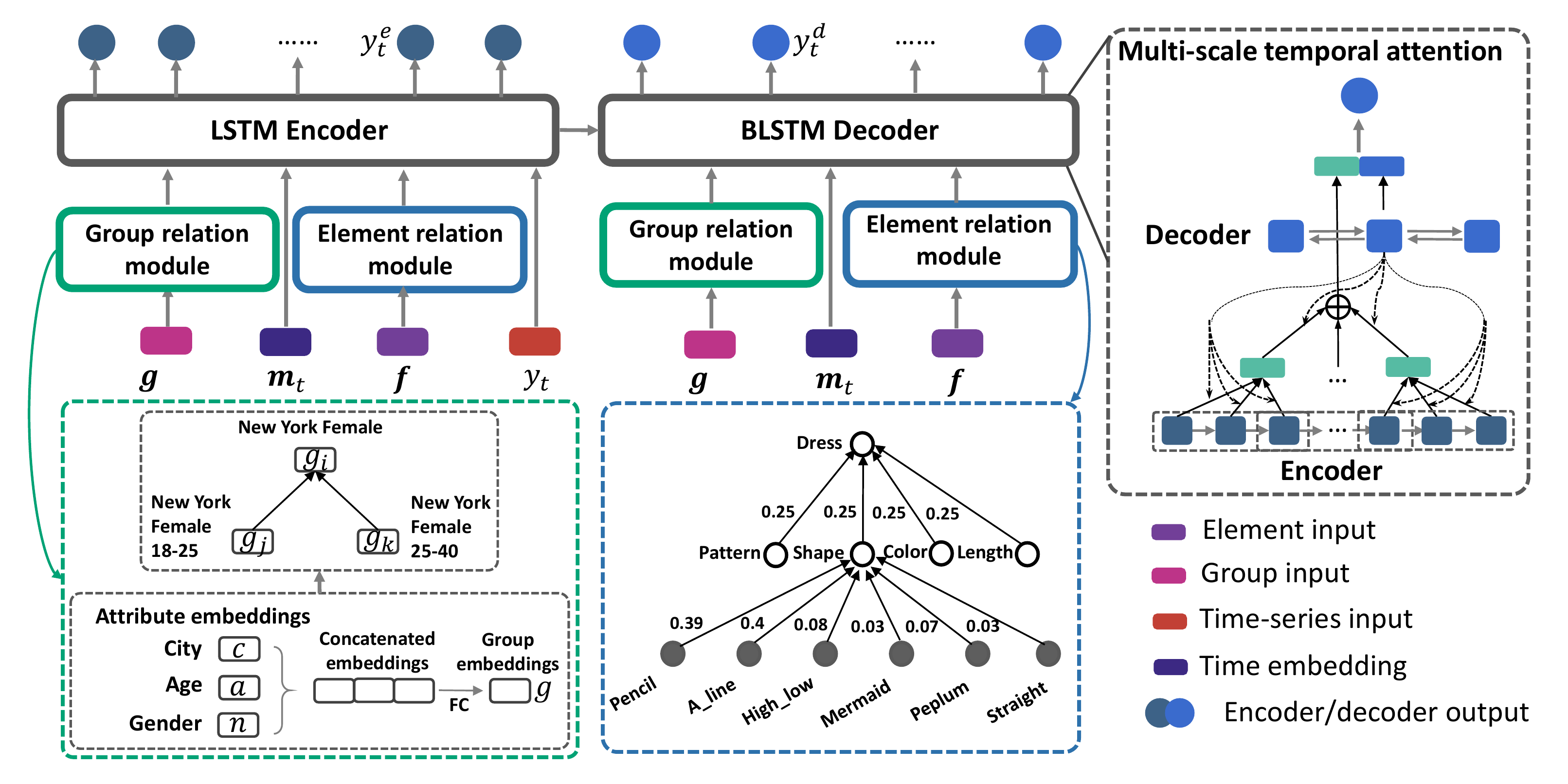}
	\caption{The REAR framework. It is based on the LSTM encoder-decoder framework and incorporates relations among groups and fashion elements (shown in the green and blue dashed line boxes respectively). Moreover, a sliding temporal attention module (shown in the black dashed line box) is used at the decoding stage to enhance the long-horizon forecasting.}
	\label{Fig:overall_framework}
\end{figure*}

\section{Approach}
This paper aims to develop an end-to-end model to forecast the fashion trends given the historical input. We adopt the LSTM encoder-decoder as the basic framework, which is able to incorporate both time series inputs and the associated sequence information into a unified model and make multi-horizon forecasting. To leverage the multiple relations among sequences, we utilize the  message passing mechanism to propagate information among related sequences thus obtain relation-enhanced representations for each sequence. Moreover, to improve the forecasting performance in long-horizon scenario, we adopt a sliding temporal attention mechanism to adaptively attend to the historical sequences, thereby countering the error accumulation effects during decoding. We name our proposed framework as \textbf{R}elation \textbf{E}nhanced \textbf{A}ttention \textbf{R}ecurrent network (REAR), as shown in Fig.~\ref{Fig:overall_framework}. In this section, we first introduce thingREAR methoing and then discuss the correlation and difference between REAR and the KERN method~\cite{ma2020knowledge}.

\subsection{Relation Enhanced Historical Trend Encoding}
\subsubsection{Sequence Feature Embedding}~\label{seq_feat_emb}
Given a time series $(y_1, \cdots, y_T)$ indicating the past trend of fashion element $f$ for group $g$ within time period $[1,T]$, we aim to forecast the future values of the trend $(y_{T+1}, \cdots, y_{T+T'})$. The group $g$ is defined by the combination of three attributes $g=[c, a, n]$, where $c \in \mathcal{C}$ is the city, $a \in \mathcal{A}$ is the age group ($\mathcal{C}$, $\mathcal{A}$ denote all cities and all age groups) and $n \in \{male, female\}$ is gender. To obtain the group representation, we first convert each of the group features $c$, $a$, and $n$ into embeddings (randomly initialized) $\pmb{c}\in \mathbb{R}^D$, $\pmb{a} \in \mathbb{R}^D$, and $\pmb{n} \in \mathbb{R}^D$ separately, where $D$ is the dimensionality of sequence feature embedding. We then adopt a linear layer to aggregate the three embeddings into one unified group representation:
\begin{equation}
    \pmb{g} = \pmb{W}_g[\pmb{c}, \pmb{a}, \pmb{n}] + \pmb{b}_g
\end{equation}
where $\pmb{W}_g \in \mathbb{R}^{D\times 3D}$, $\pmb{b}_g \in \mathbb{R}^D$, and $\pmb{g}\in \mathbb{R}^D$. For each fashion element $f$, we directly convert it into an embedding $\pmb{f} \in \mathbb{R}^D$.

\subsubsection{Multiple Relation Modeling}
As we discussed in the Introduction, fashion trend signals are not independent but correlated each other via multiple relations. Such time series correlations have also been emphasized in other time series modeling tasks such as stock price prediction~\cite{feng2019temporal} and water and air quality monitoring~\cite{liang2018geoman}. 
In terms of fashion trend, various complex relations exist, for example, the lead-lag influential patterns across different big cities~\cite{al2020paris}. However, such complex relations are difficult to specify and explore. In this paper, we start our relation incorporation from the direct, intuitive but important relations among fashion trends: the affiliation relations determined by sequence features, \textit{i.e.}, the group and fashion element.

\textbf{Relations between fashion elements}. Each piece of time series data of fashion trend describes the specific trend of one fashion element, such as \textit{dress shape: A\_line}. There are three types of fashion elements considered in this study, namely category, attribute, and style\footnote{we use the Visenze taxonomy in this study, the whole taxonomy can be found in \url{https://drive.google.com/file/d/1KgREBFE38D9U2T1uKrCrm2vRrTWuI3Po/view?usp=sharing}}. Two of them (category and attribute) can be naturally organized with a tree-structured taxonomy. As part of the taxonomy shown in the blue dashed box in Fig.~\ref{Fig:overall_framework}, the category \textit{Dress} has several attributes (\textit{e.g.}, \textit{Pattern, Shape, Color}) and the attribute \textit{Shape} has several values (\textit{e.g.}, \textit{Pencil, A\_line, High\_low}). 
It is easy to understand that because many attributes are affiliated to certain categories, the corresponding attribute values (\textit{i.e.}, \textit{A\_line}) are produced fully or partially based on one category (\textit{i.e.}, \textit{dress}). This indicates that the trends of the affiliated fashion attributes can reflect the trend of the related category to some extent. 
Considering such relations between categories and attributes,  we design to let the child nodes affect the parent nodes.

The normalized proportion of each child node out of its parent node is denoted in the edge of the taxonomy in Fig.~\ref{Fig:overall_framework}. Apparently, the trends of parent nodes will be consistent with the sum of the children nodes. For example, if we find that the trend of the attribute value \textit{peplum} goes up, it is highly probable that the category \textit{dress} also goes up. Note that this correlation of time series trends is directed, which only holds from children to parents and does not hold vice versa. For example, if the number of \textit{Shape} goes up, the number of \textit{Pencil} does not definitely go up because it may be caused by the increasing of other attribute values such as \textit{A\_line}.

To model such affiliation relations, we apply the message passing mechanism. Specifically, we have three types of nodes in this tree: \textit{category}, \textit{attribute}, and \textit{attribute value}, and the affiliation relations are between \textit{attribute} and \textit{category}, \textit{attribute value} and \textit{attribute}. As mentioned in Section~\ref{seq_feat_emb}, each fashion element $f$ is converted into a vector representation $\pmb{f}$. And the message passing is conducted between those embeddings, \textit{i.e.}, passing messages from child nodes to their parent nodes. The message passed to node $i$ is: 
\begin{equation}
  \pmb{s}_i = \sum_{j\in{\mathcal{E}_i}}\alpha^i_j {\pmb{f}_j},
\end{equation}
where $\alpha^i_{j}$ is the weight which is proportional to the impact of element $j$ on element $i$. During the implementation, we set 
$\alpha^i_{j}$ to the normalized proportion of each child node \textit{w.r.t} its parent node.
$\mathcal{E}_i$ denotes the set of fashion elements that affect $i$, \textit{i.e.}, elements affiliated to $i$. Therefore, $\pmb{s}_i \in \mathbb{R}^D$ is the message passed from all affiliated child nodes. We finally aggregate the propagated information and the original node representation using a linear layer and therefore generate the relation enhanced fashion element embeddings.  
\begin{equation}
\pmb{f}^*_i = \pmb{W}_e [\pmb{f}_i, \pmb{s}_i] + \pmb{b}_e. 
\label{eq3}
\end{equation}
where $[,]$ denotes the concatenation operation of two vectors, $\pmb{W}_e \in \mathbb{R}^{D\times 2D}$ and $\pmb{b}_e \in \mathbb{R}^D$ are the parameters of this linear layer.

\textbf{Relations between groups}. Each time series data describes the fashion trend of a certain user group, which is defined by two required attributes (city and gender) and one optional attribute (age group). The groups with the attribute of \textit{age group} are naturally more fine-grained than their corresponding groups without the attribute (but with same \textit{city} and \textit{gender}). The fine-grained groups can be treated as affiliations of their corresponding coarse-grained groups. As shown in Fig.~\ref{Fig:task}, there are three groups [\textit{New York Female 25-40 years old}], [\textit{New York Female 18-25 years old}] and [\textit{New York Female}]. Apparently, the two fine-grained groups [\textit{New York Female 25-40 years old}] and [\textit{New York Female 18-25 years old}] are affiliations (subsets) of the coarse-grained group [\textit{New York Female}]. We conduct message passing between group embeddings to leverage such relations, similar as those between fashion elements:
\begin{equation}
\left\{
\begin{aligned}
&\pmb{r}_i = \sum_{j\in{\mathcal{G}_i}}\beta^i_j {\pmb{g}_j} \\
&\pmb{g}^*_i = \pmb{W}_g [\pmb{g}_i, \pmb{r}_i] + \pmb{b}_g. 
\end{aligned}
\right.
\label{eq4}
\end{equation}
$\pmb{r}_i$ is the propagated information from group $i's$ affiliated groups $\mathcal{G}_i$, $\beta^i_j$ is the message passing weight between two groups, which is initiated to 1 for all affiliation relations and 0 else. $\pmb{W}_g$ is the trainable parameter.

Note that the fashion element and user group relations are integrated with the same message passing mechanism in our approach. In our above description, we use $i$ to denote the one specific node, which can be either a fashion element or user group when considering different relations, so as $j$. The subscript $e$ and $g$ in Eq.~\ref{eq3} and Eq.~\ref{eq4} are used to indicate the corresponding parameters for specific relations, i.e., ``element" and ``group".

\subsubsection{LSTM Encoder}
We adopt the LSTM to encode the historical fashion trend sequence as well as the enhanced sequence features. Specifically, we concatenate the group representation $\pmb{g}^*$, the fashion element representation $\pmb{f}^*$, the timestep feature $\pmb{m}_t$ (the position of each point within one year, converted to vector representation thus $\pmb{m}_t\in \mathbb{R}^D$), and the trend value $y_t$ as the input of the encoder network at timestep $t$:
\begin{equation}
    \pmb{v}_t^e = [\pmb{g}^*, \pmb{f}^*, \pmb{m}_t, y_t]
    \label{eq_seq_emb}
\end{equation}
where $\pmb{v}_t^e \in \mathbb{R}^{3D+1}$. The output of the encoder LSTM is the hidden representations for the input sequence at timestep $t$, denoted as:
\begin{equation}
    \pmb{h}_t^e=LSTM^e(\pmb{v}_t^e;\pmb{h}_{t-1}^e)
\end{equation}
where $\pmb{h}_{t-1}^e, \pmb{h}_t^e \in \mathbb{R}^H$, and $H$ is the size of the hidden state. $\pmb{h}_{t-1}^e$ is the encoder hidden state at timestep $t-1$, 

\subsection{Attended Future Trend Decoding}
\subsubsection{Bi-directional LSTM Decoder}
The decoder network is a bi-directional LSTM (BiLSTM), of which the initial hidden state is $\pmb{h}_T^e$, \textit{i.e.}, the last hidden state of the encoder. At each decoding step, it takes into the input features and outputs the forecasting value. The input feature of the decoder network at timestep $t$ is $\pmb{v}_t^d = [\pmb{g}^*, \pmb{f}^*, \pmb{m}_t]$, which is different from $\pmb{v}_t^e$ by removing the trend value $y_t$ and thus $\pmb{v}_t^d \in \mathbb{R}^{3D}$. The BiLSTM can propagate information from both forward and backward directions, and the final prediction should be made after the information propagation of BiLSTM~\cite{fan2019multi}. Formally, we denote the hidden state from forward LSTM as $\overrightarrow{\pmb{h}_t^d}$ and from backward as $\overleftarrow{\pmb{h}_t^d}$. We can obtain the final hidden state $\pmb{h}_t^d$ by concatenating them as follows:
\begin{equation}
\left\{
\begin{aligned}
\overrightarrow{\pmb{h}_t^d} &= \overrightarrow{LSTM^d}(\pmb{v}_t^d; \overrightarrow{\pmb{h}_{t-1}^d}) \\
\overleftarrow{\pmb{h}_t^d} &= \overleftarrow{LSTM^d}(\pmb{v}_t^d; 
\overleftarrow{\pmb{h}_{t+1}^d}) \\
\pmb{h}_t^d &= [\overrightarrow{\pmb{h}_t^d}, \overleftarrow{\pmb{h}_t^d}]
\end{aligned}
\right.
\end{equation}
where $\overrightarrow{\pmb{h}_t^d}, \overrightarrow{\pmb{h}_{t-1}^d}, \overleftarrow{\pmb{h}_t^d}, \overleftarrow{\pmb{h}_{t+1}^d} \in \mathbb{R}^H$, and $\pmb{h}_t^d \in \mathbb{R}^{2H}$. 

\subsubsection{Sliding Temporal Attention}
Even though the LSTM encoder-decoder framework is able to model time series data, its performance will deteriorate when the sequence length increases due to the memory update mechanism. Various temporal attention mechanisms have been tried before~\cite{cinar2017position, fan2019multi} to address this problem. For example, Cinar \textit{et al.} proposed the position-based attention model~\cite{cinar2017position} over the entire history to capture pseudo-periods in the history. However, this model can be significantly diluted when applied over a long history. To counter this problem, Fan \textit{et al.}~\cite{fan2019multi} proposed to use the hidden state in each decoding step to attend to different parts of the history data and utilized a multimodal fusion scheme to combine the attended results. The purposes of separating the whole history into parts are two-folds: (1) better attention scores will be learned on shorter sequences; (2) each part of the history mimics the period in the sequence such as business cycles (one month or one quarter)~\cite{fan2019multi}. However, since there are no overlaps between any two adjacent parts, some important time spans may be inappropriately separated into two attention parts. 
To address this problem, we propose a sliding attention scheme, which performs attention over the parts of the history generated by a sliding window.

Specifically, at the decoding stage, we set a window with a fixed length (shorter than the history length) and then slide this window over the encoding history with a specific sliding step, therefore generating a list of sub-sequences. We use $\pmb{h}_i^m$ to denote the $i$-th hidden state in the $m$-th encoder sub-sequence. When calculating the $t$-th decoder step, the temporal attention weights of $\pmb{h}_i^m$ is computed by:
\begin{equation}
    p^t_{mi} = \pmb{v}^T_p \text{tanh}(\pmb{W}_p\pmb{h}^d_t + \pmb{V}_p\pmb{h}^m_i + \pmb{b}_p), 
\end{equation}
\begin{equation}
    \gamma^t_{mi} = \frac{\text{exp}(p^t_{mi})}{\sum^{T_a}_{j=1}{\text{exp}(p^t_{mj})}},
\end{equation}
where $T_a$ is the length of the temporal attention window. Then, the attended content vectors $\pmb{c}_t$ and the transformed $\pmb{d}_t$ of sub-sequence $m$ are:
\begin{equation}
    \pmb{c}^t_m = \sum^{T_a}_{t_{i=1}}{\gamma^t_{mi}\pmb{h}^m_i}
\end{equation}
\begin{equation}
    \pmb{d}^t_m = \text{ReLU}(\pmb{W}_d\pmb{c}^t_m+\pmb{b}_d) 
\end{equation}
As shown in the right-top of Fig.~\ref{Fig:overall_framework}, we apply temporal attention on each sliding window of the encoded history and then fuse them with the multimodal attention~\cite{fan2019multi} with weights $\phi^{t}_{1...M}$, which is specifically obtained as follows:
\begin{equation}
    q^{t}_{m} = \pmb{v}^T_q \text{tanh}(\pmb{W}_q\pmb{h}^{d}_t + \pmb{V}_q\pmb{d}^{t}_m + \pmb{b}_q), 
\end{equation}
\begin{equation}
    \phi^{t}_m = \frac{\text{exp}(q^t_m)}{\sum^{M}_{k=1}{\text{exp}(q^t_k)}}.
\end{equation}
Note that $M$ indicates the number of sub-sequences generated by sliding window, which is determined by $T_a$ and the sliding step $l$.
Finally, the overall information obtained by the sliding attention from the history encoding is:
\begin{equation}
    \pmb{x}_t = \sum^{M}_{m=1}\phi^{t}_m\pmb{d}^{t}_m.
\end{equation}
We then concatenate the hidden states of the BiLSTM decoder and the attention information thus obtain the final enhanced hidden state at each decoding step:
\begin{equation}
    \pmb{h}_t^{d*} = \pmb{W}_x[\pmb{h}_t^d, \pmb{x}_t] + \pmb{b}_x.
\end{equation}
All $\pmb{W}$, $\pmb{V}$ are corresponding weight matrices with proper dimensions and all $\pmb{b}$ are the bias.

Different settings of window length and sliding step could make difference in the performance and computational complexity of the REAR method. Specifically, the window length determines the information of each attended historical part and the sliding step determines the density of the attention operation over the whole historical series. The smaller the sliding step is, the better performance it might achieve as the historical trends are attended densely, but at the same time, it will result in higher computational cost. To achieve comprehensively best performance, the two parameters should be carefully tuned and chosen. 

\subsection{Model training}
The prediction is made based on the final hidden state at each step, 
for both encoding and decoding stages during training. However, for testing, predictions only happen at the decoding stage. Particularly, we apply linear layers to make predictions in encoder and decoder respectively:
\begin{equation}
\left\{
\begin{aligned}
y_t^e &= \pmb{W}_e\pmb{h}_t^e + b_e \\
y_t^d &= \pmb{W}_d\pmb{h}_t^{d*} + b_d
\end{aligned}
\right.
\end{equation}
where $\pmb{W}_e, \pmb{W}_d \in \mathbb{R}^{1\times 2H}$ and $b_e, b_d \in \mathbb{R}$ are the parameters of the linear layer; $y_t^e, y_t^d \in \mathbb{R}$ are the forecasting value at each time step for encoder and decoder respectively. We use L1 loss to train the whole model, including the encoder loss $L_e(\cdot)$ and decoder loss $L_d(\cdot)$:
\begin{equation}
    L = L_e(\pmb{y}_e, \pmb{y}_e^{\ast}, \pmb{\theta}_e) + L_d(\pmb{y}_d, \pmb{y}_d^{\ast}, \pmb{\theta}_d)
\end{equation}
where $\pmb{\theta}_e$, $\pmb{\theta}_d$ are the model parameters for the encoder and decoder respectively; $\pmb{y}_e, \pmb{y}_e^{\ast} \in \mathbb{R}^{(T-1)}$ are the prediction and ground-truth of the encoder sequence; and $\pmb{y}_d, \pmb{y}_d^{\ast} \in \mathbb{R}^{T'}$ are the prediction and ground-truth of the decoder sequence.

\subsection{Discussion on the comparison between REAR and KERN}
REAR is developed based on KERN, both of them are based on the LSTM encoder-decoder framework and try to leverage relations between fashion trends to enhance the fashion trend modeling and forecasting. However, REAR is improved compared to KERN in terms of two main points. First, rather than only leveraging the relations between fashion elements, REAR also considers the relations between user groups, which is leveraged by conducting message passing on related user groups. Second, to improve the forecasting performance in long-horizon scenario, in the decoding stage, a sliding temporal attention mechanism is adopted to adaptively attend to the historical sequences, thereby countering the error accumulation effects during decoding. In the following experiments part, the two parts of technical improvement will be further evaluated and discussed in detail. 
\section{Experiments}
To verify the effectiveness of our proposed model, we conducted extensive experiments on two datasets. In particular, we are interested in the following research questions:
\begin{itemize}
\item[$\bullet$] \textbf{RQ1}: Can the REAR model make better fashion trend forecasting compared with the current state-of-the-art models in terms of forecasting accuracy?
\item[$\bullet$] \textbf{RQ2}: Can the introduced modules of incorporating multiple relations and sliding attention mechanism improve the performance?
\item[$\bullet$] \textbf{RQ3}: How does the REAR model perform in trend forecasting in terms of specific fashion elements, and based on that, how can the model produce insightful fashion trend forecasting?
\end{itemize}
\subsection{Experimental Settings}
\textbf{Experimental Setup}. We evaluate our model on two fashion trend forecasting datasets, our proposed FIT dataset and the GeoStyle dataset~\cite{mall2019geostyle}. Each time series in GeoStyle spans over three years and each week has one data point, while FIT dataset spans over five years and has one data point for every half month. Because of the different timespans and granularities of the two datasets, we apply different schema to form the data samples. For GeoStyle, we take one year's historical time series as input and forecast the following half year's trends. For FIT, we take two year's historical time series as input and forecast the future trends with two settings: the following nine months and twelve months. The sliding window strategy is applied to both datasets to generate the aforementioned data samples. As the total length of fashion trends in GeoStyle is shorter than that of FIT, we keep the last sample of each full sequence as the testing sample on GeoStyle, while keep the last 6 samples of each full sequence as the testing samples on FIT. Compared with \cite{ma2020knowledge} which only takes the last sample of each full sequence for testing on FIT, we increase the number of test samples to make the test results more convincing and less biased.
Note that the parts for prediction in all testing samples will not be included in training process. We use Mean Absolute Error (MAE) and Mean Absolute Percentage Error (MAPE) as the evaluation metrics~\cite{mall2019geostyle} as majority time-series prediction works do, which are specifically calculated as follows:
\begin{equation}
    MAE = \frac{1}{m}\sum_{i=1}^m\|y^*_i-y_i\|
\end{equation}
\begin{equation}
    MAPE = \frac{1}{m}\sum_{i=1}^m\|\frac{y^*_i-y_i}{y^*_i}\|
\end{equation}

\begin{table}[!t]
\caption{Performance of the REAR and baseline models for fashion trend Forecasting (the lower is better)}
\begin{center}
\setlength{\tabcolsep}{2.3mm}{\begin{tabular}{{ccc|cc|cc}}
\hline
Dataset &\multicolumn{2}{c|}{GeoStyle} &\multicolumn{4}{c}{FIT}\\
\cline{1-7} 
 ~\multirow{2}*{\text{Method}} &\multicolumn{2}{c|}{Six months}  &\multicolumn{2}{c|}{Nine months}  &\multicolumn{2}{c}{Twelve months}\\
~ &MAE &MAPE  &MAE  &MAPE  &MAE  &MAPE  \\
\cline{1-7} 
    \text{Mean} &0.0292 &25.79  &0.1131  &51.50 &0.1205  &54.99   \\
    \text{Last} &0.0226 &21.04  &0.1531  &62.20  &0.1392  &52.85  \\
    \text{AR} &0.0211	&20.69	&0.1143	&45.44	&0.1186	&46.69 \\
    \text{VAR} &0.0150 &17.95	&0.1042	&45.29	&0.1021	&41.41 \\
    \text{ES} &0.0228	&20.59	&0.1531	&62.20 &0.1392	&52.85\\
    \text{Linear} &0.0365 &24.40 &0.1749 &58.11 &0.1948 &64.56 \\ 
    \text{Cyclic} &0.0165	&16.64	&0.1626	&56.76	&0.1746	&60.65\\
    \text{GeoStyle} &0.0149	&16.03	&0.1582	&54.93	&0.1735	&60.20  \\
    \text{MM-ATT} &0.0137	&15.31	&0.0898	&32.79	&0.0976 &37.02 \\
    \text{KERN} &0.0134	&\textbf{14.24} &0.0886 &30.50 &0.1020 &34.64 \\ 
    \textbf{REAR} &\textbf{0.0134}  &14.36 &\textbf{0.0864}	&\textbf{29.45}	&\textbf{0.0951}	&\textbf{32.26}  \\
\cline{1-7} 
\cline{1-7} 

\hline
\end{tabular}}
\label{tab:overall experiment}
\end{center}
\end{table}

\textbf{Implementation Details}. We set the embedding size of user attributes (including the embeddings of city, age, and gender), fashion elements and timestep to 10 and the hidden state size of both encoder and decoder LSTM network to 50. Since each sequence in GeoStyle only has one attribute (city) and does not have any user attributes of age and gender, we simplify the fusion of group attributes and just use the city embedding as its group embedding. 
The hyper parameters $T_a$ is selected from \{12, 24\} on both datasets and $l$ is selected from  \{1, 2, 5, 10, 20\} on GeoStyle and \{1, 2, 4, 8, 12, 24\} on FIT. $T_a$ and $l$ are finally set to $[12, 20]$, $[24, 2]$, $[24, 4]$ respectively for three experimental settings: [GeoStyle, 6-month prediction], [FIT, 9-month prediction], and [FIT, 12-month prediction]. 

Compared with the statistical models like AR, VAR, and ES, which learn one model for each time series, deep learning-based models learn a unified model for all time series in the dataset. Thus, the varying magnitudes of different time series greatly affect the training of the deep learning-based models. To counter this problem, we apply the min-max normalization on both datasets. For GeoStyle, to have a fair comparison with the previous methods which is based on the dataset without any normalization, we first conduct the min-max normalization in the preprocessing of data (for all deep learning-based methods), and then convert the predicted values to original magnitudes using the pre-saved min and max values (\textit{i.e.}, a min-max de-normalization operation) when obtaining the prediction results. For FIT, we conduct the min-max normalization on the entire dataset and all the training and prediction for both baselines and our model are based on the normalized dataset.

During training, we randomly sample a batch of 400 different time series for each iteration. For each experimental setting, we train one REAR model for all fashion elements. For the evaluation of REAR model, we curate a validation set by further separating the testing data: we use the data points with odd indexes as the test set, and those with even indexes as the validation set. To make the evaluation result more robust and convincing (to avoid the case that superb performance is achieved by chance), we evaluate the model every epoch when it starts to converge. For each evaluation, we calculate the average validation results of the last ten evaluation steps. When the average validation result achieves the best, we take the corresponding average testing result as the final performance. Note that the evaluation strategy applied in this paper is different as that in \cite{ma2020knowledge}, which picks the best performance from one epoch. Comparatively, the evaluation in \cite{ma2020knowledge} is less representative as the performance may fluctuate and be unstable.

\textbf{Baseline Methods}. We compare the REAR model with several methods:
\begin{itemize}
    \item \textbf{Mean} and \textbf{Last}: They use the mean value or the value of last point of the input historical data as the forecasting value. 
    \item \textbf{Autoregression (AR)}: It is a linear regressor which uses the linear combination of last few observed values as the forecasting value.
    \item \textbf{Vector Autoregression (VAR)}: It is another stochastic process model which generalizes the univariate AR model by allowing for more than one evolving variable.
    \item \textbf{Exponential Smoothing (ES)}~\cite{al2017fashion}: It aggregates all the historical values with an exponential decayed weight, the more recent values have higher impact on the future's forecast. 
    \item \textbf{Linear} and \textbf{Cyclic}~\cite{mall2019geostyle}: They are linear or cyclical parametric models which let historical values to fit the specific predefined model. 
    \item \textbf{Geostyle}~\cite{mall2019geostyle}: It is a parametric model combining a linear component and a cyclical component. It is the state-of-the-art fashion trend forecasting method on Geostyle dataset.
    \item \textbf{MM-ATT}~\cite{fan2019multi}: An LSTM-based encoder-decoder framework, which utilizes multimodal attention during decoding. It achieved the state-of-the-art performance for multi-horizon time-series forecasting in several tasks not related to fashion.
    \item \textbf{KERN}~\cite{ma2020knowledge}: the state-of-the-art RNN-based fashion trend forecasting method which leverages the relations between fashion elements as knowledge to connect the fashion trend signals and thus enhance the forecasting performance.  
\end{itemize}

\subsection{Overall Performance on Fashion Trend Forecasting (RQ1)}
We first evaluate our REAR model for fashion trend forecasting by comparing its performance with several classic and state-of-the-art baselines. The overall results are shown in Table~\ref{tab:overall experiment}. Based on the results, we have the following analysis and discussion: 

(1) \textbf{Discussion on the results of GeoStyle}. Three LSTM-based methods, namely MM-ATT, KERN, REAR, achieve comparable performances, which are better than other statistic model-based baselines. Specifically, KERN achieves the best MAE performance and REAR achieves the best MAPE performance. The difference in performance of various methods is not quite significant. The possible reasons for such results are manifold. First, GeoStyle is an easier fashion trend dataset as most fashion trend signals in it are highly seasonal and relatively simple. As a result, traditional statistical models such as Cyclic can also achieve preferable prediction accuracy. Second, the prediction length (six months) is comparatively short, which makes the trend forecasting less difficult. Moreover, the strength of the REAR method is in that it leverages multiple relations to enhance the individual fashion trend forecasting. In the GeoStyle, there are no group relations nor element relations to exploit, which therefore limits the performance of REAR method. 

(2) \textbf{Discussion on the results of FIT}. The proposed REAR model yields the best performance on both datasets under both evaluation metrics, especially on the FIT dataset. To be noted, it clearly outperforms the other two LSTM-based methods in terms of both the MAE results and MAPE results, which demonstrates the effectiveness of REAR method with its technical improvements when tackling more challenging fashion trend forecasting tasks. Moreover, all three LSTM-based methods outperform the statistical model-based methods by a large margin, showing the effectiveness of LSTM in modeling the complicated time-series patterns for the long-period forecasting. 
\begin{figure}[!t]
	\centering
	\includegraphics[scale = 0.4]{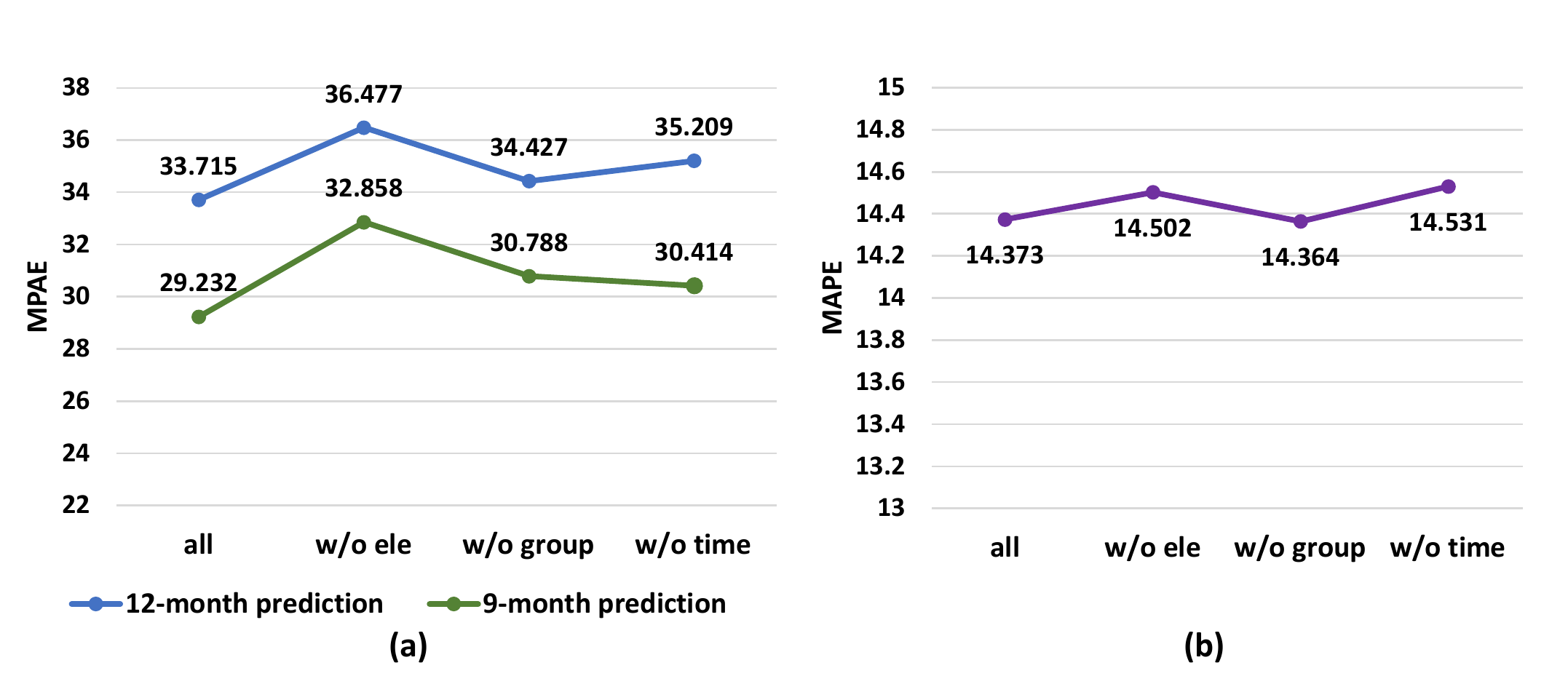}
	\vspace{-0.15in}
	\caption{Performance of the REAR model w/o three types of embeddings (MAPE result, the lower the better) on two datasets, (a) FIT and (b) GeoStyle.
	}
	\label{Fig:ablation-emb}
	\vspace{-0.15in}
\end{figure}

(3) \textbf{Discussion on different experimental settings.} Overall, most methods perform better for nine-month prediction than one-year prediction on the FIT dataset, including our REAR model. Such results are reasonable because of two reasons. First, the one-year prediction requires the model to forecast data with longer time horizon, which is apparently more difficult. Second, such setting reduces the quantity of training data. It is notable that for the nine-month prediction, the best methods are REAR and KERN while for the one-year prediction, MM-ATT is one of the best two methods. The result that MM-ATT outperforms KERN for one-year prediction demonstrates the importance of the attention mechanism for long-period forecasting. 

(4) \textbf{Comparison between REAR and KERN.} On GeoStyle, REAR achieves similar performance. As explained above, GeoStyle is an easier dataset compared with FIT for having simple fashion trend patterns and shorter forecasting period. As a result, both of the two methods achieve desired performance. Plus, there are no group relations nor element relations on GeoStyle, resulting in most technical parts in REAR not applicable. Thus, the performance of REAR is greatly limited. 

On FIT, our REAR method significantly outperforms the KERN. Recall that the technical differences of REAR compared to KERN are three-fold: leveraging more group relations, removing internal similarity relation incorporation, and introducing the sliding temporal attention mechanism. The better performance of REAR on FIT demonstrates the effectiveness of all technical improvements of REAR. More specifically, on FIT, group relations are available to incorporate. Also, as the forecasting period is comparatively long, the devised attention mechanism benefits the model more significantly on FIT. 

\begin{figure}[!t]
	\centering
	\includegraphics[scale = 0.4]{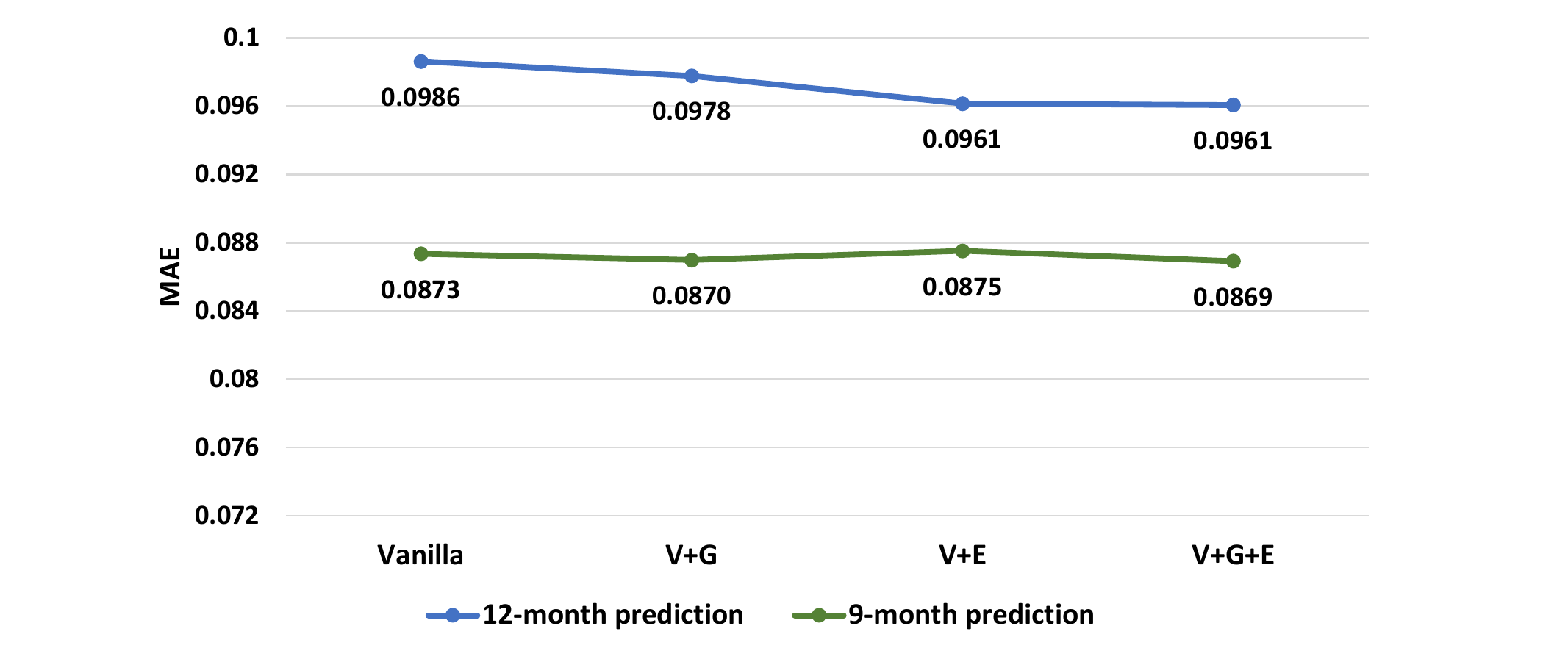}
	\vspace{-0.15in}
	\caption{Performance of the model w/o leveraging relations (MAE result, the lower the better). \textbf{V} denotes vanilla, \textbf{G} denotes group relation, \textbf{E} denote element relation. Results are on FIT dataset.   
	}
	\label{Fig:ablation-relation}
	\vspace{-0.15in}
\end{figure}

\begin{table}
  \caption{MAE performance of models with temporal attention in different sliding steps. Results on FIT dataset.}
  \begin{center}
  \label{tab:attention}
  \setlength{\tabcolsep}{4.2mm}{
  \begin{tabular}{ccc}
    \hline
    \cline{1-3} 
    sliding steps &Nine months  &Twelve months \\
    \cline{1-3} 
    \text{No Att} &0.08692 &0.09606 \\
    \text{2} &\textbf{0.08644} &0.09532\\
    \text{4}&0.08646 &\textbf{0.09514}\\
    \text{8}&0.08674 &0.09571\\
    \text{12}&0.09685 &0.09712\\
    \text{24} &0.08742 &0.09615\\
    \hline
  \end{tabular}}
  \end{center}
\end{table}

Based on the comprehensive comparison between the results of REAR and KERN, we can reach the conclusion that the proposed REAR is better for handling longer-period fashion trend forecasting tasks in which more external relations between fashion trends can be incorporated. Note that the experimental results of KERN are not exactly same as those reported in~\cite{ma2020knowledge} because of the difference in experimental setting, specifically the evaluation setting and data splitting.

\begin{figure}[!t]
	\centering
	\includegraphics[scale = 0.52]{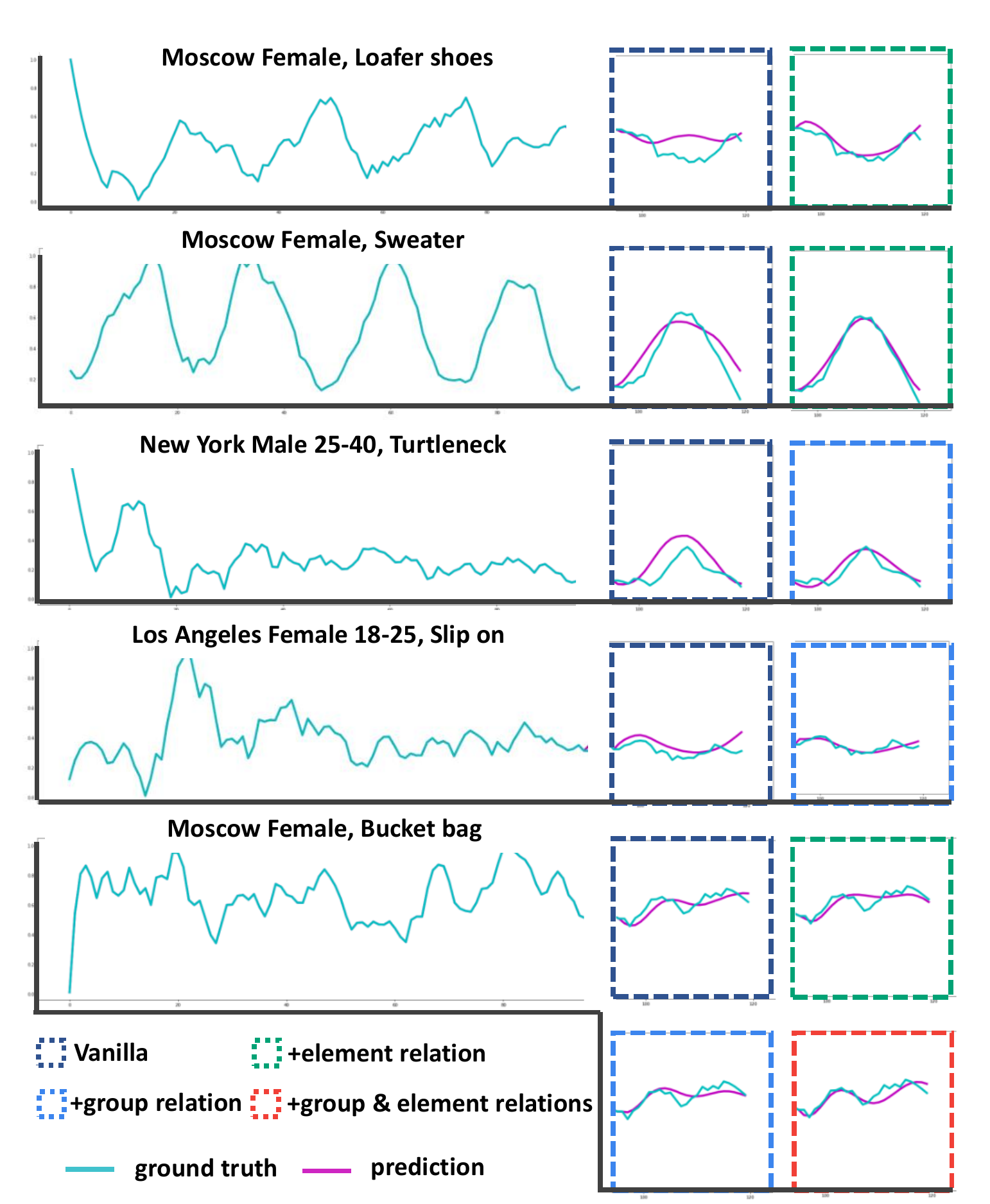}
	\vspace{-0.15in}
	\caption{Fashion trend forecasting results of models w/o two kinds of relations. In each curve figure, the X-axis is the time (one for half month) and Y-axis is the popularity index of the corresponding fashion element.
	}
	\label{Fig:pred-visual}
\end{figure}

\subsection{Ablation Study (RQ2)}
In the ablation study, we conduct several groups of experiments to validate the detailed components in the proposed model. We first evaluate the effectiveness of the three types of content embeddings used as the input in the sequential model, \textit{i.e.}, time embedding, group embedding and element embedding (explained in Eq~\ref{eq_seq_emb}). The MAPE results of four models for three experimental settings on two datasets are illustrated in Fig.~\ref{Fig:ablation-emb}. \textbf{all} denotes the model with all three types of embeddings in the basic LSTM encoder-decoder framework, while the remaining three models, \textit{i.e.}, \textbf{w/o ele}, \textbf{w/o grp}, \textbf{w/o time}, represent removing the element embedding, group embedding and time embedding respectively. The experimental results on FIT show that removing any sequence feature embedding can degrade the overall performance, demonstrating the effectiveness of all three kinds of sequence features. On GeoStyle, we can discover that both element and time embeddings are effective in improving the performance, while the group embedding is not. Because the group information in GeoStyle is limited, only with the city information, it is not enough to help the model. It is also notable that overall, the element embedding is more significant for the fashion trend modeling compared with the other two. It is probably because that the fashion element is still the deterministic factor for fashion trending. 

We further discuss the effectiveness of incorporating multiple relations in the proposed fashion trend forecasting model. Specifically, two types of relations, fashion element relations and user group relations, are exploited. We compare the experimental results of models with/without each type of relations and show the results in Fig.~\ref{Fig:ablation-relation}. The MAE results of \textbf{V+G} and \textbf{V+E} are better than that of \textbf{Vanilla} (no relation leveraged) for the 12-month fashion trend prediction. Such improvement is not quite marked in the 9-month prediction case. However, it is clear that leveraging both relations effectively improves the prediction performance. This demonstrates the effectiveness of introducing relations in helping the sequence modeling of fashion trend signals and the trend prediction. In addition, the more significant improvement for the 12-month prediction concludes that the relation knowledge is particularly helpful for more challenging longer-horion prediction cases.

\begin{figure}[!t]
	\centering
	\includegraphics[scale = 0.55]{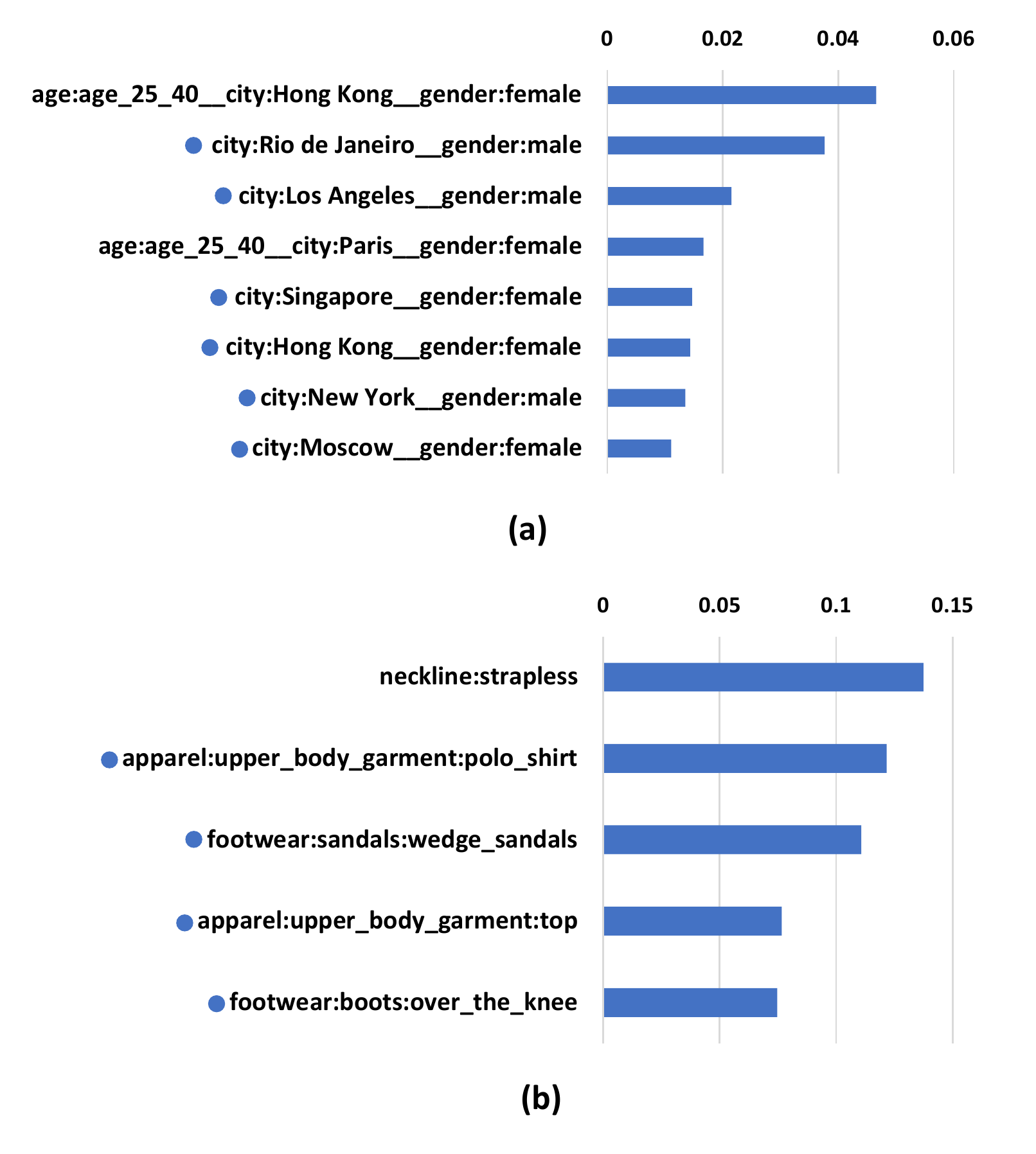}
	\vspace{-0.15in}
	\caption{Performance improvement (relative improvement on MAE) by applying relations on specific groups and fashion elements. (a) Top 8 groups in terms of performance improvement by leveraging group relations; (b) Top 5 fashion elements in terms of performance improvement by leveraging element relations. Results on 9-month prediction on FIT dataset.
	}
	\label{Fig:performance-analysis}
	\vspace{-0.1in}
\end{figure}

\begin{figure*}[!t]
	\centering
	\includegraphics[scale = 0.58]{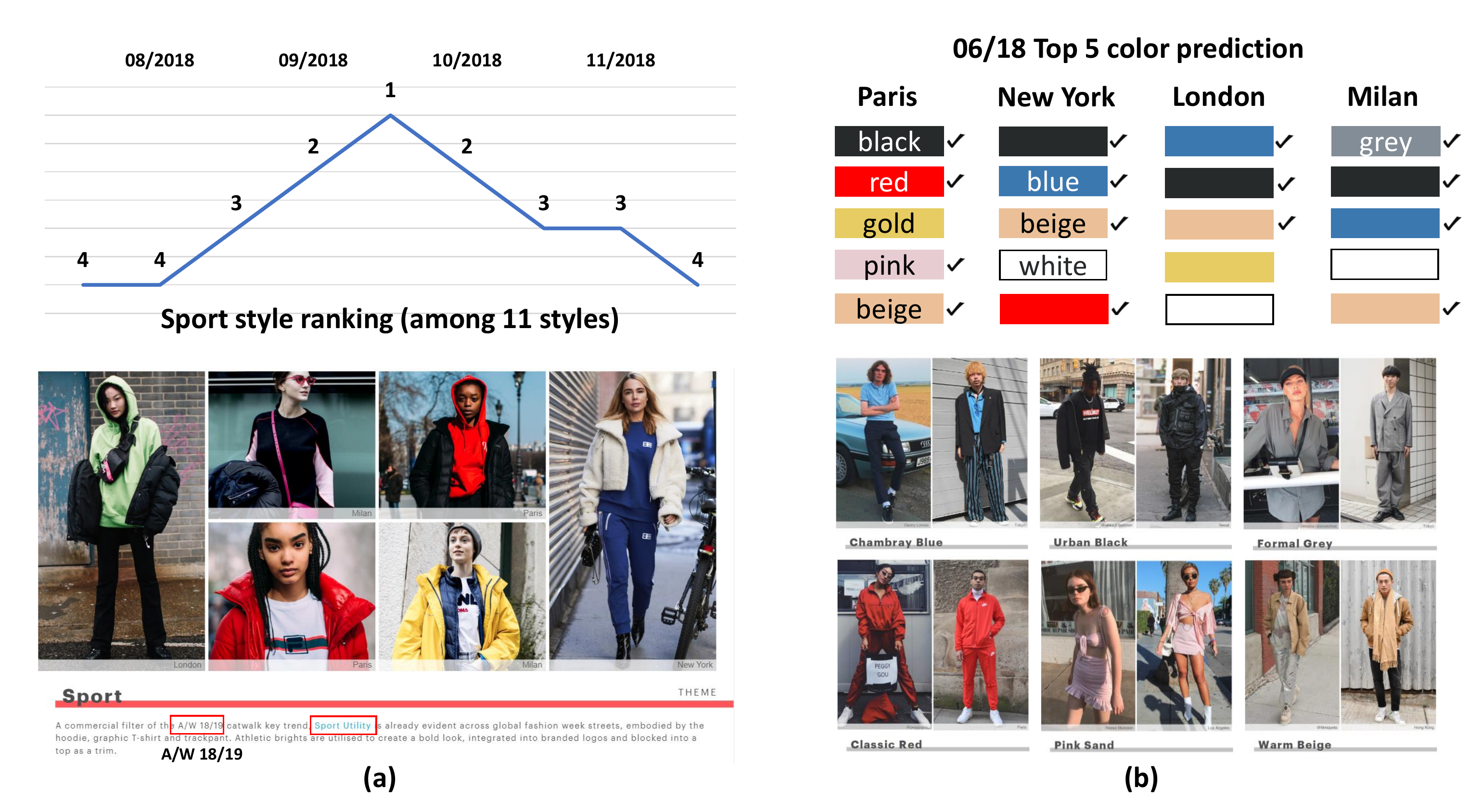}
	\vspace{-0.1in}
	\caption{Two cases of specific fashion trend predictions. (a) trend prediction of the fashion style \textbf{sporty}. top: predicted ranking of \textbf{sporty} by REAR model; bottom: professional trendy style prediction produced by WSGN for A/W 18/19. (b) color trend prediction. top: predicted top 5 colors among four major fashion metropolis in June 2018; bottom: professional trendy color prediction produced by WSGN for S/S 2018.\protect\footnotemark[5]
	}
	\label{Fig:report}
	\vspace{-0.1in}
\end{figure*}

We further discuss in detail the sliding temporal attention mechanism in the REAR model. Specifically, we analyse the effectiveness of employing the attention mechanism and the effect of varying sliding steps on the performance of the whole model. In Table~\ref{tab:attention}, we report the performance of the models (MAE results)  with varying sliding steps of the temporal attention on two experimental settings on the FIT dataset. From the results we can see that first, introducing the temporal attention with sliding steps clearly reduces the MAE for both the 9-month prediction and 12-month prediction in most cases. The difference is that when the sliding step is two, the model performs best for the 9-month prediction, while for the 12-month prediction, the best setting is four sliding steps. For both prediction settings, the performance worsens when the sliding step becomes large. However, another noteworthy point is that the small sliding step causes high computational cost, which should be considered in comprehensive model evaluation.  

\subsection{Fashion Trend Analysis (RQ3)}
In this part we show some visualization results of the fashion trend prediction, as well as some in-depth performance analysis to further discuss the effectiveness of the REAR model, especially the effectiveness of multiple relations incorporation in the model. In Fig.~\ref{Fig:pred-visual}, we illustrate five examples which are fashion trends of different fashion elements for different groups. To depict the whole picture of the data, we show the entire fashion trend signals and highlight the prediction part in the colored square in the right part. The first two rows compare the prediction results of \textbf{Vanilla} model and model leveraging element relations, from which it is clear that taking advantage of element relations makes the prediction of future trend more precisely. In the third and fourth cases, we can see the difference of prediction results brought by leveraging group relations. Similar as the first two cases, we can see that the \textbf{Vanilla} model is not good enough to make solid trend predictions, while the models incorporating relations perform better. In the last case, we show all four prediction results from models without relations, which validates the effectiveness of our assumptions about incorporating multiple relations to boost the fashion trend prediction performance. Such observations from the visualization results are consistent with the results of quantitative analysis in the ablation study part.

We further analyze how the incorporation of two kinds of relations impact the fashion trend prediction in detail. Specifically, we investigate the performance change of fashion trends belonging to \textbf{different groups} by and modeling the \textbf{group relations}. We show the top eight groups whose fashion trend prediction is improved the most in Fig.~\ref{Fig:performance-analysis} (a).
From the result we can see that the most significantly improved groups with the help of group relation incorporation are those of coarse-grained (such as Rio de Janeiro Female, without age group). Such a result is reasonable as in our model, the coarse-grained groups adsorb the information from their affiliating groups, therefore the representation of these groups are enhanced. We also investigate the impact of \textbf{element relations} on trends of \textbf{different fashion elements} and have found the similar results. In Fig.~\ref{Fig:performance-analysis} (b), we illustrate the five most improved fashion elements in terms of fashion trend prediction. As we can see, most of these fashion elements belong to the higher-level node in the taxonomy (\textit{e.g.}, category). Recall that we incorporate the relations, between groups or elements, by effectively passing information from the fined-grained affiliation nodes to their parent nodes. In practice, the parent nodes are exactly those wider covering groups and higher-level fashion elements. As discussed above, these parent nodes  adsorb  more  information, and thus are better modeled to achieve more preferable performance.  

Our REAR model is able to make fashion trend predictions regarding to specific fashion elements for a wide range of time. To evaluate the quality of the predicted trends by REAR, in Fig.~\ref{Fig:report}, we show two cases of fashion trend prediction from our REAR model and compare them with the corresponding trends made by the professional fashion forecasting agency WGSN\footnote{WGSN is one of the most acknowledged professional fashion forecasting agency. We obtain the WGSN forecasting results based on the fashion forecasting reports from the WSGN website, https://www.wgsn.com/fashion/}. Note that the trend forecasting from WGSN is expert-based while our forecasting is data-based. The chart on the top in Fig.~\ref{Fig:report} (a) is the ranking result of the style \textbf{sporty} based on the prediction result of REAR, from which we can see that the \textbf{sporty style} is predicted to become more popular since September 2018. Such forecasting results show consistency with WGSN, which forecasts that the \textbf{sport} style is the key trend in the A/W\footnote{A/W means the Fall/Winter season, which starts around late July and ends in December. S/S means the Spring/Summer season, which usually goes from January to June each year.} 18/19 season. Fig.~\ref{Fig:report} (b) shows the color forecasting results. Specifically, on the top we show the top five popular colors predicted by the REAR model in June 2018 for four major fashion cities, \textit{i.e.}, Paris, New York, London and Milan. We can see that based on our model, the most trendy colors in June 2018 include black, red, blue, and beige, grey and pink are also popular in some cities. Meanwhile, WGSN produces the S/S 18 street color trend, in which we can see most of them are in our predicted top popular color list (matching predictions are marked by ticks). From the two cases, we can see that our REAR model can not only predict the future trend, but also generate the forecasting results that are similar to the old-fashion expert-based fashion forecasting results produced by professional fashion forecasting agencies. 

\section{Conclusion}
This paper addressed the fashion trend forecasting problem based on social media, aiming to mine the complex patterns in the historical time series records of fashion elements and accordingly predict the future trends. Specifically, over 170 fashion elements were studied and three types of user information were involved towards forecasting insightful fashion trends for specific user groups. 
An effective model, Relation Enhanced Attention Recurrent network (REAR) was proposed to leverage multiple relations between specific fashion trends and therefore captured the complex patterns in the time-series data and effectively forecast fashion trends.

Although much effort had been made and desirable results had been achieved, some aspects could be further improved in the future. First, more user information that specify the user fashion preference should be explored such as occupations or hobbies. Second, more knowledge sources that have great impacts, such as fashion magazines, bloggers and brands, should be considered in the massive fashion trend forecasting.

\bibliographystyle{IEEEtran}
\bibliography{refe}

\begin{IEEEbiography}[{\includegraphics[width=1in,height=1.25in,clip,keepaspectratio]{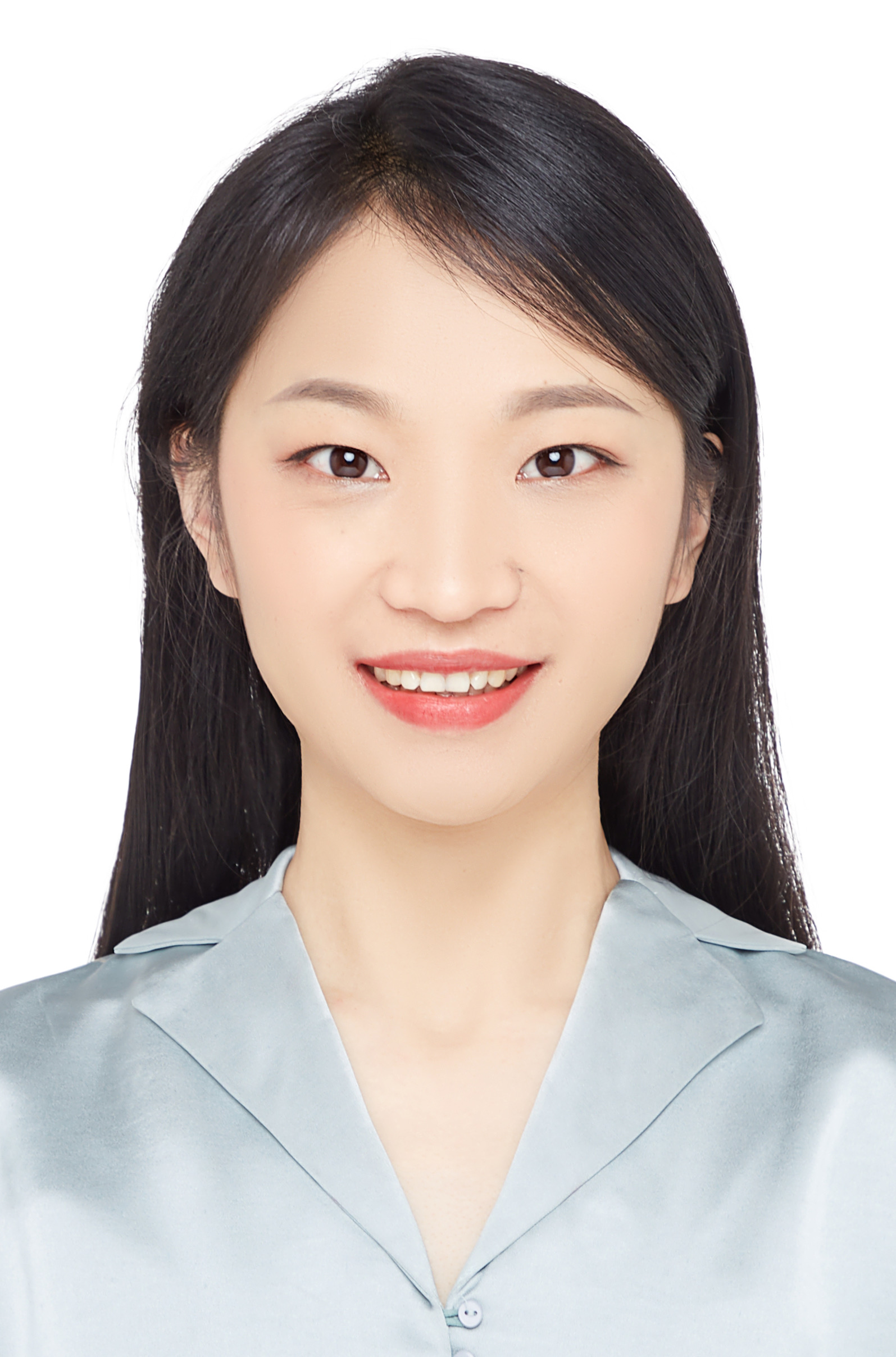}}]{Yujuan Ding} received the Ph.D. degree from The Hong Kong Polytechnic University, Hong Kong. She is currently a Research Assistant in Laboratory for Artificial Intelligence in Design, Hong Kong. Her research interests include multimedia analysis and computational fashion. She has published papers in top-tier journals such as TMM and TCSVT. She also serves as reviewer for journals including TMM, TCSVT, etc. 
\end{IEEEbiography}

\begin{IEEEbiography}[{\includegraphics[width=1in,height=1.25in,clip,keepaspectratio]{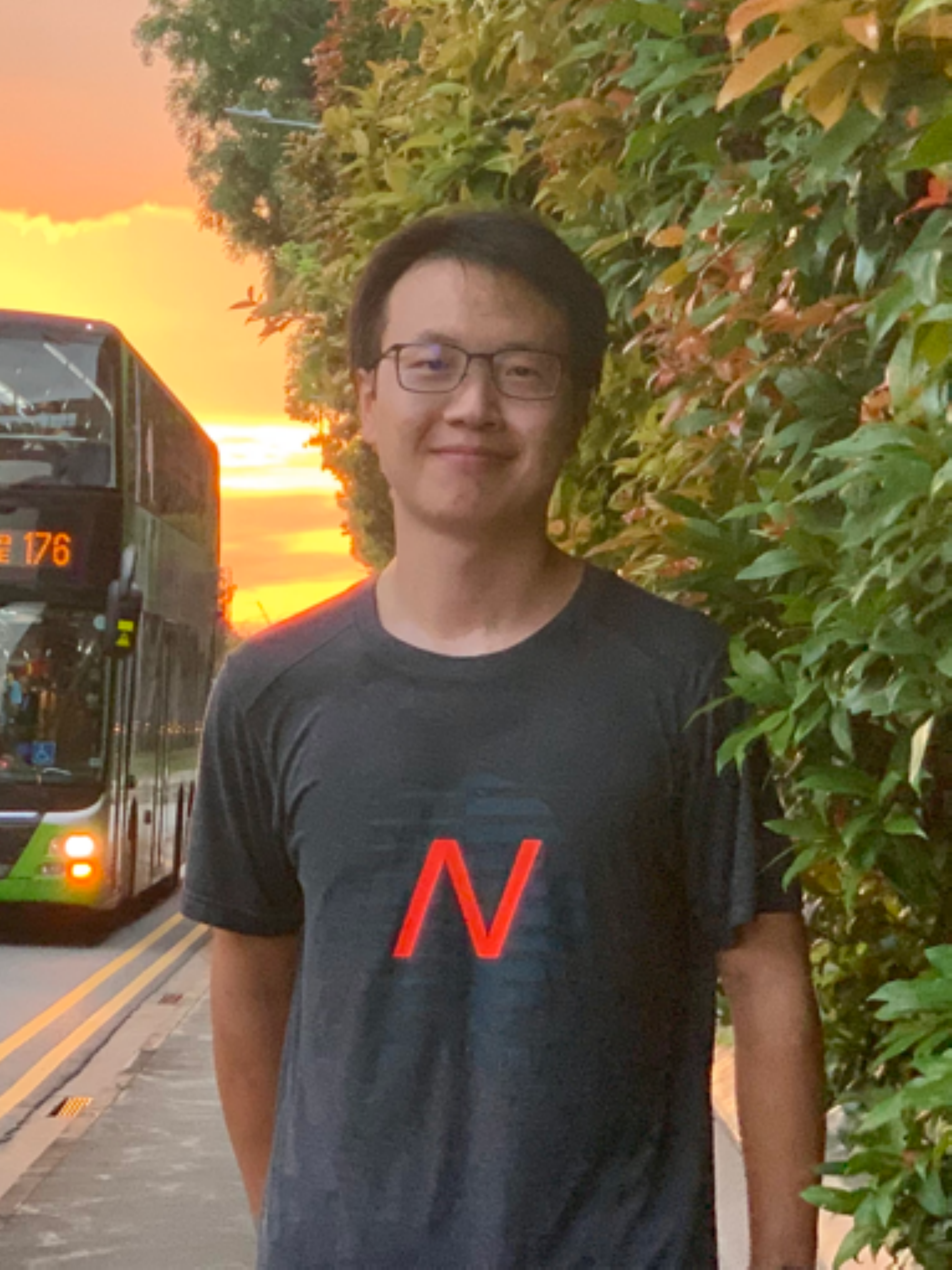}}]{Ma Yunshan} received the B.E. degree from Beijing University of Posts and Telecommunications, Beijing, China in 2015. He is currently a Ph.D. candidate of School of Computing at National University of Singapore. His research interests focus on multimedia analysis, especially for computational fashion analysis, including fashion knowledge extraction, fashion trend forecasting, and fashion recommendation.
\end{IEEEbiography}

\begin{IEEEbiography}[{\includegraphics[width=1in,height=1.25in,clip,keepaspectratio]{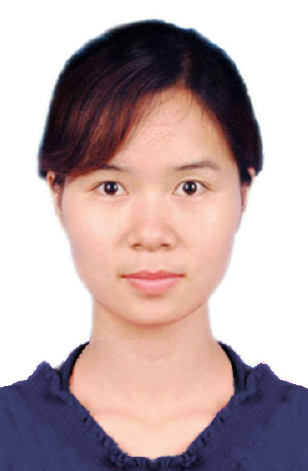}}]{Lizi Liao}
received the Ph.D. degree in 2019 from NUS Graduate School for Integrative Sciences and Engineering at
the National University of Singapore. She is currently a Research Fellow in NExT++ center, School of Computing at National University of Singapore. Her research interests include conversational system, multimedia analysis and recommendation. 
Her works have appeared in top-tier conferences such as MM, WWW, ICDE, ACL, IJCAI and AAAI, and top-tier journals such as TKDE. She received the Best Paper Award Honorable Mention of ACM MM 2018. Moreover, she has served as the PC member for international conferences including SIGIR, WSDM, ACL, and the invited reviewer for journals including TKDE, TMM and KBS.
\end{IEEEbiography}

\begin{IEEEbiography}[{\includegraphics[width=1in,height=1.25in,clip,keepaspectratio]{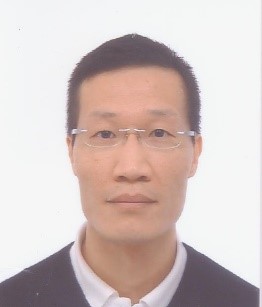}}]{Wai Keung Wong}
received the Ph.D. degree from The Hong Kong Polytechnic University, Hung Hom, Kowloon, Hong Kong. He is currently a Professor with The Hong Kong Polytechnic University. He has authored or co-authored more than 100 papers in refereed journals and conferences, including the IEEE Transactions on Neural Networks and Learning Systems, the IEEE Transactions on Systems, Man, and Cybernetics-Part B: Cybernetics and the IEEE Transactions on Systems, Man, and Cybernetics-Part C: Applications and Reviews, Pattern Recognition, CHAOS, European Journal of Operational Research, Neural Networks, Applied Soft Computing, Information Science, Decision Support Systems, etc.. His current research interests include pattern recognition and feature extraction.  
\end{IEEEbiography}

\begin{IEEEbiography}[{\includegraphics[width=1in,height=1.25in,clip,keepaspectratio]{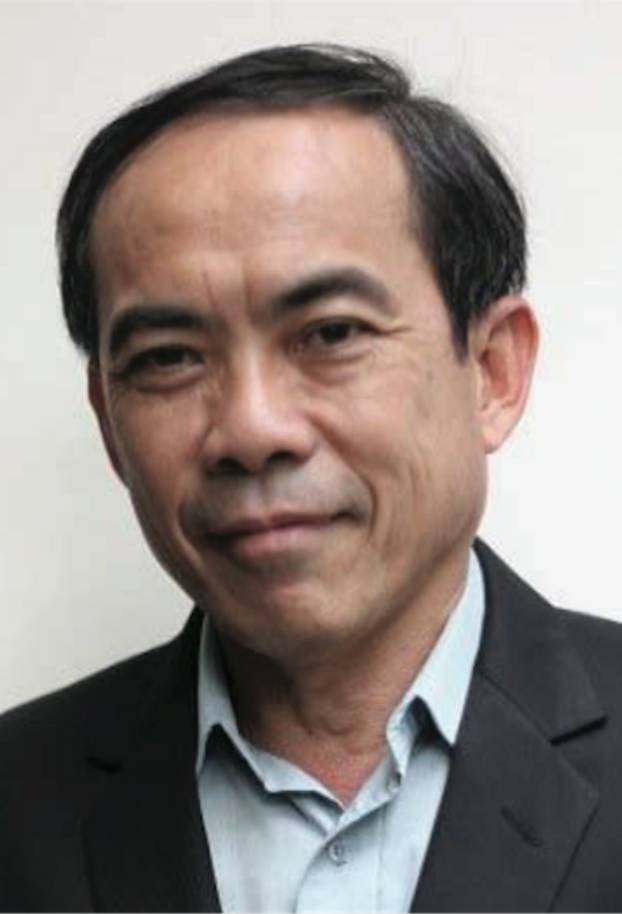}}]{Tat-Seng Chua}
is the KITHCT Chair Professor at the School of Computing, National University of Singapore. He was the Acting and Founding Dean of the School from 1998-2000. Dr Chua's main research interest is in multimedia information retrieval and social media analytics. In particular, his research focuses on the extraction, retrieval and question-answering (QA) of text and rich media arising from the Web and multiple social networks. He is the co-Director of NExT, a joint Center between NUS and Tsinghua University to develop technologies for live social media search.

Dr Chua is the 2015 winner of the prestigious ACM SIGMM award for Outstanding Technical Contributions to Multimedia Computing, Communications and Applications. He is the Chair of steering committee of ACM International Conference on Multimedia Retrieval (ICMR) and Multimedia Modeling (MMM) conference series. Dr Chua is also the General Co-Chair of ACM Multimedia 2005, ACM CIVR (now ACM ICMR) 2005, ACM SIGIR 2008, and ACM Web Science 2015. He serves in the editorial boards of four international journals. Dr. Chua is the co-Founder of two technology startup companies in Singapore. He holds a PhD from the University of Leeds, UK.
\end{IEEEbiography}

\end{document}